\documentclass[twoside,11pt]{article}

\usepackage{jmlr2e}

\usepackage{xcolor}
\definecolor{diffcolor}{rgb}{0.16, 0.32, 0.75}

\usepackage{lastpage}
\jmlrheading{22}{2021}{1-\pageref{LastPage}}{12/20; Revised
7/21}{9/21}{20-1473}{Hubert Baniecki, Wojciech Kretowicz, Piotr Piatyszek, Jakub Wisniewski, and Przemyslaw Biecek}

\ShortHeadings{dalex: Responsible Machine Learning in Python}{Baniecki, Kretowicz, Piatyszek, Wisniewski, and Biecek}
\firstpageno{1}

\begin{document}

\title{
dalex: Responsible Machine Learning with Interactive Explainability and Fairness in Python
}

\author{\name Hubert Baniecki${}^{1}$\email hubert.baniecki.stud@pw.edu.pl \\
    \name Wojciech Kretowicz${}^{1}$\email wojciech.kretowicz.stud@pw.edu.pl \\
    \name Piotr Piatyszek${}^{1}$\email piotr.piatyszek.stud@pw.edu.pl \\
    \name Jakub Wisniewski${}^{1}$\email jakub.wisniewski10.stud@pw.edu.pl \\
    \name Przemyslaw Biecek${}^{1,2}$\email przemyslaw.biecek@pw.edu.pl \\
    ${}^{1}$\addr Faculty of Mathematics and Information Science, Warsaw University of Technology, Poland\\
    ${}^{2}$\addr Samsung Research \& Development Institute, Poland
}

\editor{Joaquin Vanschoren}

\maketitle

\begin{abstract}
In modern machine learning, we observe the phenomenon of \emph{opaqueness debt}, which manifests itself by an increased risk of discrimination, lack of reproducibility, and deflated performance due to data drift. An increasing amount of available data and computing power results in the growing complexity of black-box predictive models. To manage these issues, good MLOps practice asks for better validation of model performance and fairness, higher explainability, and continuous monitoring. The necessity for deeper model transparency comes from both scientific and social domains and is also caused by emerging laws and regulations on artificial intelligence. To facilitate the responsible development of machine learning models, we introduce \texttt{dalex}, a Python package which implements a model-agnostic \emph{interface} for interactive explainability and fairness. It adopts the design crafted through the development of various tools for explainable machine learning; thus, it aims at the \emph{unification} of existing solutions. This library's source code and documentation are available under open license at \url{https://python.drwhy.ai}.
\end{abstract}

\begin{keywords}
  explainability, fairness, interactivity, interpretability, responsible AI
\end{keywords}

\section{Introduction}

From the evolution of statistical modeling through data mining and machine learning to so-called artificial intelligence (AI), we arrived at the point where advanced systems support, or even surpass, humans in various predictive tasks. These algorithms are available for broad user-bases through numerous machine learning frameworks in Python like \texttt{scikit-learn} \citep{scikit-learn}, \texttt{tensorflow} \citep{tensorflow}, \texttt{xgboost} \citep{xgboost} or \texttt{lightgbm} \citep{lightgbm} to name just a few. Nowadays, there are increased concerns regarding the explainability \citep{Lipton2018, Miller2019} and fairness \citep{fairness1, fairness2} of machine learning predictive models in research and commercial domains. A growing number of stakeholders discuss various needs and features for frameworks related to responsible machine learning \citep{responsible-ai, responsible-ml}. For us, the primary objective is combining three aspects of model analysis: explainability, fairness, and crucially for human-model dialogue, interactivity~\citep{interactive-xai}. 

Related software most notably include Python packages from these three categories. \texttt{lime} \citep{lime}, \texttt{shap} \citep{shap}, \texttt{pdpbox} \citep{pdpbox}, \texttt{interpret} \citep{interpretml}, \texttt{alibi} \citep{alibi}, and \texttt{aix360} \citep{aix360} implement various explainability methods; \texttt{aif360} \citep{aif360}, \texttt{aequitas} \citep{aequitas}, and \texttt{fairlearn} \citep{fairlearn} implement various fairness methods; moreover, responsible AI tools for \texttt{tensorflow} \citep{tensorflow}, e.g. \texttt{witwidget} \citep{whatiftool}, produce interactive dashboards supporting machine learning operations (this is also partially addressed by \texttt{interpret} and \texttt{fairlearn}). All these leave room for improvement in terms of the combining of various methods, while also connecting them to ever-growing modeling and data frameworks through a uniform abstraction layer. 

Unlike many of the proposed solutions, we strongly emphasize the construction of end-to-end software for facilitating a responsible approach to machine learning. To achieve that, we focus on tabular data while there are frameworks specializing in other modalities, e.g. \texttt{innvestigate} \citep{innvestigate}. The \texttt{dalex} package unifies various approaches and bridges the existing gap separating black-box models from explainability methods. Moreover, \texttt{dalex} brings numerous fairness metrics and interactive model analysis dashboards closer to the user. These factors motivate our article, in which we preview our previous work in Section \ref{sec:previous}, introduce \texttt{dalex} in Section \ref{sec:package}, and sketch the future work in Section~\ref{sec:conclusion}. 

\section{Previous Work}
\label{sec:previous}

This contribution builds upon the software for explainable machine learning presented by us in \textit{``DALEX: Explainers for Complex Predictive Models in R''} \citep{dalex}. Since \texttt{DALEX} version \texttt{0.2.5}, there have been two major releases, which expanded the toolkit of explainability methods, and performed a complete redesign of code, interface and charts for model visualizations. Users provided us with a number of very valuable feature requests: (i) we created a~taxonomy of model-agnostic explanations for machine learning predictive models \citep{emabook}; (ii) we prototyped \texttt{modelStudio} \citep{modelStudio}, an extension of \texttt{DALEX}, which automatically produces a customizable dashboard allowing for an interactive model analysis \citep{iema}; (iii) we added support for multi-output predictive models and a growing number of machine learning frameworks in a language-agnostic manner. Further, we noticed that the visual model analysis goes beyond the area of explainability and also addresses such issues as fairness and interactive model comparisons. Based on these experiences, we implemented a Python package.

\section{A Unified Interface for Responsible Machine Learning}
\label{sec:package}

\begin{figure}[!t]
    \centering
    \includegraphics[width=\linewidth]{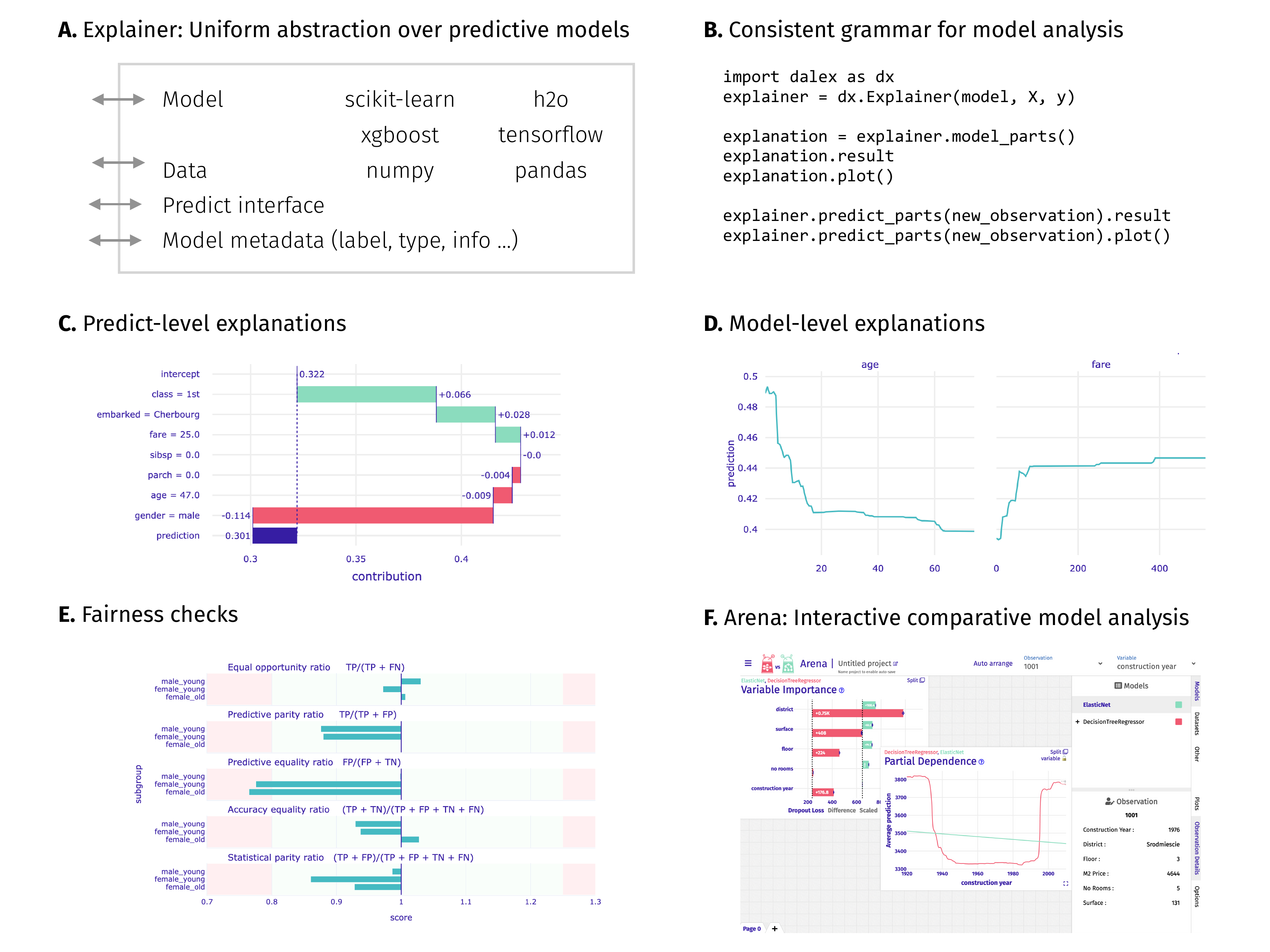}
    \caption{The \texttt{dalex} package is based on six pillars that support responsible machine learning modeling: \textbf{A.} The main \texttt{Explainer} class object, which serves as a uniform abstraction over predictive models and data API's in Python; \textbf{B.} A unified set of methods for model analysis with explanation objects that calculate results and plot them in a consistent way; \textbf{C.} Predict-level (local) explainability methods; \textbf{D.} Model-level (global) explainability methods; \textbf{E.} Fairness oriented methods; \textbf{F.} Interactive dashboard for comparative model analysis.}
    \vspace{-0.5cm}
    \label{fig:interface}
\end{figure}

The \texttt{dalex} Python package implements the main \texttt{dalex.Explainer} class to provide an abstract layer between distinct model API's (e.g. \texttt{scikit-learn} \citep{scikit-learn}, \texttt{tensorflow} \citep{tensorflow}, \texttt{xgboost} \citep{xgboost}, \texttt{h2o} \citep{h2o}) and data API's (e.g. \texttt{numpy} \citep{numpy}, \texttt{pandas} \citep{pandas}), and the explainability and fairness methods. In Figure \ref{fig:interface}, we present the architecture of a~unified interface for model-agnostic responsible machine learning with interactive explainability and fairness. These methods are divided into model-level techniques operating on a whole dataset (or~its subset) and predict-level techniques operating on distinct observations from data (or~their neighbourhoods). The binding of these methods to the one \texttt{dalex.Explainer} class gives a favourable user experience, where one can conveniently compute and return various explanation objects. All of them share the main \texttt{result} attribute, which is a~\texttt{pandas.DataFrame}, and the \texttt{plot} method, which produces visualizations with the \texttt{plotly} package \citep{plotly}. The latter takes multiple explanation objects, which allows for an easy model comparison.

\paragraph{Model-level and predict-level explanations.}

Explainability methods referenced in Figure \ref{fig:interface} return different objects depending on the \texttt{type} parameter: \texttt{model\_performance} and \texttt{predict} allow for easy interference with the model basics, \texttt{predict\_parts} implements iBreakDown local variable attributions and Shapley values estimation, \texttt{model\_parts} implements permutational variable importance, \texttt{predict\_profile} implements Ceteris Paribus profiles, \texttt{model\_profile} implements PDP, ALE and ICE profiles, \texttt{model\_diagnostics} implements overall diagnostics of models' residuals, \texttt{model\_surrogate} implements surrogate decision tree models, which are effective to \texttt{plot}. Additionally, the \texttt{dalex.Explainer} abstract layer allows for the integration of other explanations, e.g. the \texttt{shap} \citep{shap} explanations into \texttt{predict\_parts} and \texttt{model\_parts} methods, and \texttt{lime} \citep{lime} into \texttt{predict\_surrogate}. All of these methods are described in detail in the \textit{EMA} book \citep{emabook} with \texttt{dalex} Python code examples.

\paragraph{Fairness checks.}

The principles of responsible machine learning involve providing proper model accountability and bias detection  \citep{responsible-ai, responsible-ml}. Because of regulations and guidelines, we can see an increasing demand for easily accessible methods to check model fairness \citep{fairness1, fairness2}. Therefore, we implemented the \texttt{fairness\_check} method, which compares the most common fairness measures based on the confusion matrix \citep{bias, metrics} and provides a detailed textual description of the group fairness analysis. It operates on a fairness object available through the \texttt{dalex.Explainer.model\_fairness} method. In the same way as explanation objects, it contains the \texttt{result} attribute and \texttt{plot} method, which provides various visualizations depending on the \texttt{type} parameter. 

\paragraph{Interactive and comparative model analysis.}

The user-centred design of explainable (responsible) AI tools brings other emerging challenges discussed on the junction of AI and HCI domains \citep{interactive-xai, Miller2019}. The \texttt{dalex.Arena} class creates an advanced live \texttt{Arena} dashboard \citep{arena} for model comparisons with all features available in the \texttt{dalex} package, including model explainability and fairness, moreover techniques for data exploration. These allow the juxtaposition of various visualizations for model and data analysis, which gives a complete view of the various models' behaviour. Notably, the dashboard can be saved into a local state to be loaded later --- this overcomes the reproducibility crisis apparent in machine learning.

\section{Conclusion and Future Work}
\label{sec:conclusion}

In this article, we present \texttt{dalex}, which builds upon and extends the \texttt{DALEX} R package to bring a unified interface for responsible machine learning into Python. This package is continuously developed, while the current stable version \texttt{1.3} for Python \texttt{3.9} is available at \url{https://python.drwhy.ai}. Due to the comprehensive design of a uniform abstraction layer, \texttt{dalex} allows for the convenient addition of new machine learning frameworks into the responsible realm, which is not the case for most of the existing solutions. Additionally, with a clear-cut taxonomy of methods, there is the possibility to add new explanation objects and metrics, which was well-proven within our previous work. We further discuss such matters in the documentation and educational materials attached to this package.

We next aim to include into \texttt{dalex} explanations for groups of interacting variables, which is a highly influential concept in modern machine learning algorithms. There is research to be done towards adding a \texttt{predict\_fairness} method, as the individual fairness field is not that well established. Overall, the responsible machine learning domain aims to address more principles than explainability and fairness \citep{responsible-ai}; thus, the next steps shall address the accountability, robustness, and safety of machine learning models.

\clearpage

\acks{We want to thank the \url{https://drwhy.ai} community, users, researchers, and developers, for the continuous valid feedback over the past years. This work was financially supported by the \texttt{(Poland) NCN Opus grant 2017/27/B/ST6/0130}.}

\vskip 0.2in
\bibliography{references.bib}

\begin{thebibliography}{33}
\providecommand{\natexlab}[1]{#1}
\providecommand{\url}[1]{\texttt{#1}}
\expandafter\ifx\csname urlstyle\endcsname\relax
  \providecommand{\doi}[1]{doi: #1}\else
  \providecommand{\doi}{doi: \begingroup \urlstyle{rm}\Url}\fi

\bibitem[Abadi et~al.(2016)Abadi, Barham, Chen, Chen, Davis, Dean, Devin,
  Ghemawat, Irving, Isard, Kudlur, Levenberg, Monga, Moore, Murray, Steiner,
  Tucker, Vasudevan, Warden, Wicke, Yu, and Zheng]{tensorflow}
Mart\'{\i}n Abadi, Paul Barham, Jianmin Chen, Zhifeng Chen, Andy Davis, Jeffrey
  Dean, Matthieu Devin, Sanjay Ghemawat, Geoffrey Irving, Michael Isard,
  Manjunath Kudlur, Josh Levenberg, Rajat Monga, Sherry Moore, Derek~G. Murray,
  Benoit Steiner, Paul Tucker, Vijay Vasudevan, Pete Warden, Martin Wicke, Yuan
  Yu, and Xiaoqiang Zheng.
\newblock {TensorFlow: A System for Large-Scale Machine Learning}.
\newblock \emph{USENIX Conference on Operating Systems Design and
  Implementation}, pages 265--283, 2016.

\bibitem[Abdul et~al.(2018)Abdul, Vermeulen, Wang, Lim, and
  Kankanhalli]{interactive-xai}
Ashraf Abdul, Jo~Vermeulen, Danding Wang, Brian~Y. Lim, and Mohan Kankanhalli.
\newblock {Trends and Trajectories for Explainable, Accountable and
  Intelligible Systems: An HCI Research Agenda}.
\newblock \emph{CHI Conference on Human Factors in Computing Systems}, pages
  1--18, 2018.

\bibitem[Alber et~al.(2019)Alber, Lapuschkin, Seegerer, H{{\"a}}gele,
  Sch{{\"u}}tt, Montavon, Samek, M{{\"u}}ller, D{{\"a}}hne, and
  Kindermans]{innvestigate}
Maximilian Alber, Sebastian Lapuschkin, Philipp Seegerer, Miriam H{{\"a}}gele,
  Kristof~T. Sch{{\"u}}tt, Gr{{\'e}}goire Montavon, Wojciech Samek,
  Klaus-Robert M{{\"u}}ller, Sven D{{\"a}}hne, and Pieter-Jan Kindermans.
\newblock {iNNvestigate Neural Networks!}
\newblock \emph{Journal of Machine Learning Research}, 20\penalty0
  (93):\penalty0 1--8, 2019.

\bibitem[Arya et~al.(2020)Arya, Bellamy, Chen, Dhurandhar, Hind, Hoffman,
  Houde, Liao, Luss, Mojsilovic, Mourad, Pedemonte, Raghavendra, Richards,
  Sattigeri, Shanmugam, Singh, Varshney, Wei, and Zhang]{aix360}
Vijay Arya, Rachel K.~E. Bellamy, Pin-Yu Chen, Amit Dhurandhar, Michael Hind,
  Samuel~C. Hoffman, Stephanie Houde, Q.~Vera Liao, Ronny Luss, Aleksandra
  Mojsilovic, Sami Mourad, Pablo Pedemonte, Ramya Raghavendra, John~T.
  Richards, Prasanna Sattigeri, Karthikeyan Shanmugam, Moninder Singh, Kush~R.
  Varshney, Dennis Wei, and Yunfeng Zhang.
\newblock {AI Explainability 360: An Extensible Toolkit for Understanding Data
  and Machine Learning Models}.
\newblock \emph{Journal of Machine Learning Research}, 21\penalty0
  (130):\penalty0 1--6, 2020.

\bibitem[Baniecki and Biecek(2019)]{modelStudio}
Hubert Baniecki and Przemyslaw Biecek.
\newblock {modelStudio}: Interactive studio with explanations for {ML}
  predictive models.
\newblock \emph{Journal of Open Source Software}, 4\penalty0 (43):\penalty0
  1798, 2019.

\bibitem[Baniecki and Biecek(2020)]{iema}
Hubert Baniecki and Przemyslaw Biecek.
\newblock {The Grammar of Interactive Explanatory Model Analysis}.
\newblock \emph{arXiv preprint arXiv:2005.00497}, 2020.

\bibitem[Barredo~Arrieta et~al.(2019)Barredo~Arrieta, Diaz~Rodriguez, Del~Ser,
  Bennetot, Tabik, Barbado~González, Garcia, Gil-Lopez, Molina, Benjamins,
  Chatila, and Herrera]{responsible-ai}
Alejandro Barredo~Arrieta, Natalia Diaz~Rodriguez, Javier Del~Ser, Adrien
  Bennetot, Siham Tabik, Alberto Barbado~González, Salvador Garcia, Sergio
  Gil-Lopez, Daniel Molina, V.~Richard Benjamins, Raja Chatila, and Francisco
  Herrera.
\newblock {Explainable Artificial Intelligence (XAI): Concepts, taxonomies,
  opportunities and challenges toward responsible AI}.
\newblock \emph{Information Fusion}, 58:\penalty0 82--115, 2019.

\bibitem[Bellamy et~al.(2018)Bellamy, Dey, Hind, Hoffman, Houde, Kannan, Lohia,
  Martino, Mehta, Mojsilovic, Nagar, Ramamurthy, Richards, Saha, Sattigeri,
  Singh, Varshney, and Zhang]{aif360}
Rachel K.~E. Bellamy, Kuntal Dey, Michael Hind, Samuel~C. Hoffman, Stephanie
  Houde, Kalapriya Kannan, Pranay Lohia, Jacquelyn Martino, Sameep Mehta,
  Aleksandra Mojsilovic, Seema Nagar, Karthikeyan~Natesan Ramamurthy, John
  Richards, Diptikalyan Saha, Prasanna Sattigeri, Moninder Singh, Kush~R.
  Varshney, and Yunfeng Zhang.
\newblock {AI Fairness 360: An Extensible Toolkit for Detecting, Understanding,
  and Mitigating Unwanted Algorithmic Bias}.
\newblock \emph{arXiv preprint arXiv:1810.01943}, 2018.

\bibitem[Biecek(2018)]{dalex}
Przemyslaw Biecek.
\newblock {DALEX: Explainers for Complex Predictive Models in R}.
\newblock \emph{Journal of Machine Learning Research}, 19\penalty0
  (84):\penalty0 1--5, 2018.

\bibitem[Biecek and Burzykowski(2021)]{emabook}
Przemyslaw Biecek and Tomasz Burzykowski.
\newblock \emph{{Explanatory Model Analysis}}.
\newblock Chapman and Hall/CRC, New York, 2021.
\newblock ISBN 9780367135591.
\newblock URL \url{https://pbiecek.github.io/ema}.

\bibitem[Binns(2018)]{fairness1}
Reuben Binns.
\newblock {Fairness in Machine Learning: Lessons from Political Philosophy}.
\newblock \emph{Conference on Fairness, Accountability and Transparency},
  81:\penalty0 149--159, 2018.

\bibitem[Bird et~al.(2020)Bird, Dud{\'i}k, Edgar, Horn, Lutz, Milan, Sameki,
  Wallach, and Walker]{fairlearn}
Sarah Bird, Miro Dud{\'i}k, Richard Edgar, Brandon Horn, Roman Lutz, Vanessa
  Milan, Mehrnoosh Sameki, Hanna Wallach, and Kathleen Walker.
\newblock {Fairlearn: A toolkit for assessing and improving fairness in AI}.
\newblock Technical Report MSR-TR-2020-32, Microsoft, 2020.

\bibitem[Chen and Guestrin(2016)]{xgboost}
Tianqi Chen and Carlos Guestrin.
\newblock {XGBoost: A Scalable Tree Boosting System}.
\newblock \emph{ACM SIGKDD International Conference on Knowledge Discovery and
  Data Mining}, pages 785--794, 2016.

\bibitem[Feldman et~al.(2015)Feldman, Friedler, Moeller, Scheidegger, and
  Venkatasubramanian]{bias}
Michael Feldman, Sorelle~A. Friedler, John Moeller, Carlos Scheidegger, and
  Suresh Venkatasubramanian.
\newblock {Certifying and Removing Disparate Impact}.
\newblock \emph{ACM SIGKDD International Conference on Knowledge Discovery and
  Data Mining}, pages 259--268, 2015.

\bibitem[Gill et~al.(2020)Gill, Hall, Montgomery, and Schmidt]{responsible-ml}
Navdeep Gill, Patrick Hall, Kim Montgomery, and Nicholas Schmidt.
\newblock {A Responsible Machine Learning Workflow with Focus on Interpretable
  Models, Post-hoc Explanation, and Discrimination Testing}.
\newblock \emph{Information}, 11\penalty0 (3):\penalty0 137, 2020.

\bibitem[H2O.ai(2020)]{h2o}
H2O.ai.
\newblock \emph{{Python Interface for H2O}}, 2020.
\newblock URL \url{https://github.com/h2oai/h2o-3}.

\bibitem[Harris et~al.(2020)Harris, Millman, van~der Walt, Gommers, Virtanen,
  Cournapeau, Wieser, Taylor, Berg, Smith, Kern, Picus, Hoyer, van Kerkwijk,
  Brett, Haldane, del R{'{\i}}o, Wiebe, Peterson, G{'{e}}rard-Marchant,
  Sheppard, Reddy, Weckesser, Abbasi, Gohlke, and Oliphant]{numpy}
Charles~R. Harris, K.~Jarrod Millman, St{'{e}}fan~J. van~der Walt, Ralf
  Gommers, Pauli Virtanen, David Cournapeau, Eric Wieser, Julian Taylor,
  Sebastian Berg, Nathaniel~J. Smith, Robert Kern, Matti Picus, Stephan Hoyer,
  Marten~H. van Kerkwijk, Matthew Brett, Allan Haldane, Jaime~Fern{'{a}}ndez
  del R{'{\i}}o, Mark Wiebe, Pearu Peterson, Pierre G{'{e}}rard-Marchant, Kevin
  Sheppard, Tyler Reddy, Warren Weckesser, Hameer Abbasi, Christoph Gohlke, and
  Travis~E. Oliphant.
\newblock {Array programming with NumPy}.
\newblock \emph{Nature}, 585\penalty0 (7825):\penalty0 357--362, 2020.

\bibitem[Holstein et~al.(2019)Holstein, Wortman~Vaughan, Daum\'{e}, Dudik, and
  Wallach]{fairness2}
Kenneth Holstein, Jennifer Wortman~Vaughan, Hal Daum\'{e}, Miro Dudik, and
  Hanna Wallach.
\newblock {Improving Fairness in Machine Learning Systems: What Do Industry
  Practitioners Need?}
\newblock \emph{CHI Conference on Human Factors in Computing Systems}, pages
  1--16, 2019.

\bibitem[Jiangchun(2018)]{pdpbox}
Li~Jiangchun.
\newblock \emph{PDPbox: python partial dependence plot toolbox}, 2018.
\newblock URL \url{https://github.com/SauceCat/PDPbox}.

\bibitem[Ke et~al.(2017)Ke, Meng, Finley, Wang, Chen, Ma, Ye, and
  Liu]{lightgbm}
Guolin Ke, Qi~Meng, Thomas Finley, Taifeng Wang, Wei Chen, Weidong Ma, Qiwei
  Ye, and Tie-Yan Liu.
\newblock {LightGBM: A Highly Efficient Gradient Boosting Decision Tree}.
\newblock \emph{Advances in Neural Information Processing Systems},
  30:\penalty0 3146--3154, 2017.

\bibitem[Klaise et~al.(2021)Klaise, Looveren, Vacanti, and Coca]{alibi}
Janis Klaise, Arnaud~Van Looveren, Giovanni Vacanti, and Alexandru Coca.
\newblock {Alibi Explain: Algorithms for Explaining Machine Learning Models}.
\newblock \emph{Journal of Machine Learning Research}, 22\penalty0
  (181):\penalty0 1--7, 2021.

\bibitem[Lipton(2018)]{Lipton2018}
Zachary~C. Lipton.
\newblock {The Mythos of Model Interpretability}.
\newblock \emph{Queue}, 16\penalty0 (3):\penalty0 31--57, 2018.

\bibitem[Lundberg and Lee(2017)]{shap}
Scott~M Lundberg and Su-In Lee.
\newblock {A Unified Approach to Interpreting Model Predictions}.
\newblock \emph{Advances in Neural Information Processing Systems}, pages
  4768--4777, 2017.

\bibitem[Miller(2019)]{Miller2019}
Tim Miller.
\newblock {Explanation in artificial intelligence: Insights from the social
  sciences}.
\newblock \emph{Artificial Intelligence}, 267:\penalty0 1--38, 2019.

\bibitem[Nori et~al.(2019)Nori, Jenkins, Koch, and Caruana]{interpretml}
Harsha Nori, Samuel Jenkins, Paul Koch, and Rich Caruana.
\newblock {InterpretML: A Unified Framework for Machine Learning
  Interpretability}.
\newblock \emph{arXiv preprint arXiv:1909.09223}, 2019.

\bibitem[Parmer and Kruchten(2020)]{plotly}
Chris Parmer and Nicolas Kruchten.
\newblock \emph{{plotly: An open-source, interactive data visualization library
  for Python}}, 2020.
\newblock URL \url{https://github.com/plotly/plotly.py}.

\bibitem[Pedregosa et~al.(2011)Pedregosa, Varoquaux, Gramfort, Michel, Thirion,
  Grisel, Blondel, Prettenhofer, Weiss, Dubourg, Vanderplas, Passos,
  Cournapeau, Brucher, Perrot, and {{\'E}}douard Duchesnay]{scikit-learn}
Fabian Pedregosa, Ga{{\"e}}l Varoquaux, Alexandre Gramfort, Vincent Michel,
  Bertrand Thirion, Olivier Grisel, Mathieu Blondel, Peter Prettenhofer, Ron
  Weiss, Vincent Dubourg, Jake Vanderplas, Alexandre Passos, David Cournapeau,
  Matthieu Brucher, Matthieu Perrot, and {{\'E}}douard Duchesnay.
\newblock {Scikit-learn: Machine Learning in Python}.
\newblock \emph{Journal of Machine Learning Research}, 12\penalty0
  (85):\penalty0 2825--2830, 2011.

\bibitem[Piatyszek and Biecek(2020)]{arena}
Piotr Piatyszek and Przemyslaw Biecek.
\newblock \emph{{Arena: universal dashboard for model exploration}}, 2020.
\newblock URL \url{https://arena.drwhy.ai/}.

\bibitem[Ribeiro et~al.(2016)Ribeiro, Singh, and Guestrin]{lime}
Marco~Tulio Ribeiro, Sameer Singh, and Carlos Guestrin.
\newblock {``Why Should I Trust You?'': Explaining the Predictions of Any
  Classifier}.
\newblock \emph{ACM SIGKDD International Conference on Knowledge Discovery and
  Data Mining}, pages 1135--1144, 2016.

\bibitem[Saleiro et~al.(2018)Saleiro, Kuester, Hinkson, London, Stevens,
  Anisfeld, Rodolfa, and Ghani]{aequitas}
Pedro Saleiro, Benedict Kuester, Loren Hinkson, Jesse London, Abby Stevens, Ari
  Anisfeld, Kit~T. Rodolfa, and Rayid Ghani.
\newblock {Aequitas: A Bias and Fairness Audit Toolkit}.
\newblock \emph{arXiv preprint arXiv:1811.05577}, 2018.

\bibitem[Verma and Rubin(2018)]{metrics}
Sahil Verma and Julia Rubin.
\newblock {Fairness Definitions Explained}.
\newblock \emph{International Workshop on Software Fairness}, pages 1--7, 2018.

\bibitem[{W}es {M}c{K}inney(2010)]{pandas}
{W}es {M}c{K}inney.
\newblock {D}ata {S}tructures for {S}tatistical {C}omputing in {P}ython.
\newblock \emph{{P}ython in {S}cience {C}onference}, pages 56--61, 2010.

\bibitem[Wexler et~al.(2020)Wexler, Pushkarna, Bolukbasi, Wattenberg, Viegas,
  and Wilson]{whatiftool}
James Wexler, Mahima Pushkarna, Tolga Bolukbasi, Martin Wattenberg, Fernanda
  Viegas, and Jimbo Wilson.
\newblock {The What-If Tool: Interactive Probing of Machine Learning Models}.
\newblock \emph{IEEE Transactions on Visualization and Computer Graphics},
  26\penalty0 (1):\penalty0 56--65, 2020.

\end{thebibliography}

\end{document}